\documentclass[letterpaper]{article} % DO NOT CHANGE THIS
\usepackage{aaai25}  % DO NOT CHANGE THIS
\usepackage{times}  % DO NOT CHANGE THIS
\usepackage{helvet}  % DO NOT CHANGE THIS
\usepackage{courier}  % DO NOT CHANGE THIS
\usepackage[hyphens]{url}  % DO NOT CHANGE THIS
\usepackage{graphicx} % DO NOT CHANGE THIS
\usepackage{natbib}  % DO NOT CHANGE THIS AND DO NOT ADD ANY OPTIONS TO IT
\usepackage{caption} % DO NOT CHANGE THIS AND DO NOT ADD ANY OPTIONS TO IT
\usepackage{dsfont}
\usepackage{multirow}
\usepackage{booktabs}
\usepackage[ruled]{algorithm2e}

\usepackage{newfloat}
\usepackage{listings}
\DeclareCaptionStyle{ruled}{labelfont=normalfont,labelsep=colon,strut=off} % DO NOT CHANGE THIS
\title{Personalized Sleep Staging Leveraging Source-free \\Unsupervised Domain Adaptation}
\author{
    Yangxuan Zhou\textsuperscript{\rm 1,2}, Sha Zhao\textsuperscript{\rm 1,2\thanks{Corresponding authors}}, Jiquan Wang\textsuperscript{\rm 1,2}, Haiteng Jiang\textsuperscript{\rm 3,4,1}, \\Shijian Li\textsuperscript{\rm 1,2}, Benyan Luo\textsuperscript{\rm 5}, Tao Li\textsuperscript{\rm 3,4,1}, Gang Pan\textsuperscript{\rm 1,2,4}
}
\affiliations{
    \textsuperscript{\rm 1}State Key Laboratory of Brain-machine Intelligence, Zhejiang University, Hangzhou, China\\ \textsuperscript{\rm 2}College of Computer Science and Technology, Zhejiang University, Hangzhou, China\\ \textsuperscript{\rm 3}Department of Neurobiology, Affiliated Mental Health Center \& Hangzhou Seventh People’s Hospital, Zhejiang University School of Medicine, Hangzhou, China\\ \textsuperscript{\rm 4}MOE Frontier Science Center for Brain Science and Brain-machine Integration, Zhejiang University, Hangzhou, China \\ \textsuperscript{\rm 5}The First Affiliated Hospital, College of Medicine, Zhejiang University, Hangzhou, China \\ 
    \{zyangxuan, szhao, wangjiquan, h.jiang, shijianli, luobenyan, litaozjusc, gpan\}@zju.edu.cn
}

\usepackage{bibentry}
\begin{document}

\maketitle

\begin{abstract}
Sleep staging is important for monitoring sleep quality and diagnosing sleep-related disorders. Recently, numerous deep learning-based models have been proposed for automatic sleep staging using polysomnography recordings. Most of them are trained and tested on the same labeled datasets which results in poor generalization to unseen target domains. 
However, they regard the subjects in the target domains as a whole and overlook the individual discrepancies, which limits the model's generalization ability to new patients (i.e., unseen subjects) and plug-and-play applicability in clinics.
To address this, we propose a novel Source-Free Unsupervised Individual Domain Adaptation (SF-UIDA) framework for sleep staging, leveraging sequential cross-view contrasting and pseudo-label based fine-tuning. It is actually a two-step subject-specific adaptation scheme, which enables the source model to  effectively adapt to newly appeared unlabeled individual without access to the source data.
It meets the practical needs in real-world scenarios, where the personalized customization can be plug-and-play applied to new ones.
Our framework is applied to three classic sleep staging models and evaluated on three public sleep datasets, achieving the state-of-the-art performance.
\end{abstract}


	\section{INTRODUCTION}
Sleep plays a crucial role in people's lives and has a significant impact on their overall well-being \cite{humphreys2005we}. Sleep staging is important for monitoring sleep quality and serves as a valuable tool to help diagnose sleep disorders\cite{10242085}, which refers to classify sleep periods into different stages. Recently, Polysomnography (PSG) has been widely used for sleep staging in clinics, which records various physiological signals by the sensors attached to different parts of one body, such as electroencephalography (EEG), electrooculography (EOG), and electromyography (EMG). The PSG recordings are usually divided into consecutive epochs of 30s. Experts manually identify each epoch into five distinct sleep stages, namely, W, N1, N2, N3, REM, according to the American Academy of Sleep Medicine (AASM)~\cite{aasm}. Obviously, such process is subjective, time-consuming and labor-intensive.

Numerous deep learning-based models have been proposed for automatic sleep staging \cite{tsinalis2016automatic, mousavi2019sleepeegnet, 10504925, cbramod} in recent years. Such models are usually trained and tested on the same labeled source data, automatically classifying different stages. They have achieved good performance for sleep staging in the valid set, however, their performance is not so satisfying on unknown samples. Most of them overlook the \textbf{individual discrepancies}, such as physiological structures \cite{matsushima2012age}, physical characteristics (i.e., Electrodermal Response) and sleep habits. In clinical practice, the new patients (unknown subjects) are probably different from the samples used for training the models offline. This limits the model's generalization ability to unknown subjects and dramatically degrades the performance, especially when they are directly tested on the new patients in clinics. Therefore, we desperately need to make the model adapt to new subjects.
\begin{figure}[!tb]
	\centering
	\includegraphics[width=1.0\columnwidth]{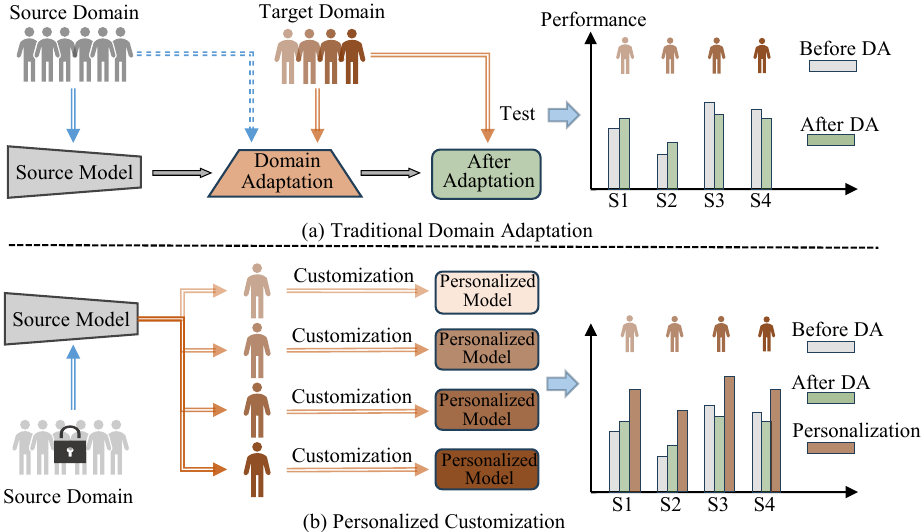}
	\caption{(a) Traditional Domain Adaptation for a group of target (new) individuals and (b) Source-free Unsupervised Personalized Customization for each target (new) individual.}
	\label{fig:0}
\end{figure}
Unsupervised Domain Adaptation (UDA) is a suitable method which can transfer knowledge learned from a labeled dataset (i.e., source domain) to an unlabeled dataset (i.e., target domain). Traditional UDA approaches commonly apply a joint training strategy, which relies on not only the target but also the source data to mitigate domain shift through feature alignment. By doing so, the performance of some target subjects is improved after adaptation (e.g.,  S1, S2 subjects in Fig. \ref{fig:0} (a)). However, this adapting way will lead to the following drawbacks. First, the target domain is treated as a whole (i.e., a number of subjects), which requires waiting until a batch of target domain samples are available before conducting the adaptation. This is impractical in real life, where the arrival of each new individual is entirely random. \textbf{We need a plug-and-play adaptation rather than waiting for all the target individuals to arrive to conduct adaptation}. Second, they overlook the individual discrepancies, resulting in a failure to adapt to certain special ones whose distribution deviates significantly from the overall distribution (e.g., S3, S4 subjects in Fig. \ref{fig:0} (a)). \textbf{A personalized sleep staging model is needed for each new patient}. Third, using source data for joint-training is time-consuming and can lead to data privacy leakage. \textbf{A source-free adaptation is needed in clinical practice.}

To meet such practical needs, in this paper, we propose an \textbf{S}ource-\textbf{F}ree \textbf{U}nsupervised \textbf{I}ndividual \textbf{D}omain \textbf{A}daptation framework for automatic sleep staging, named SF-UIDA.
First, we introduce \textbf{the concept of individual domains}, where the SF-UIDA framework treats each target subject as a distinct target domain.
Meanwhile, the proposed SF-UIDA contains a two-step \textbf{subject-specific strategy} that considers individual discrepancies to mitigate their impact on the source model. It can be applied to the source pretrained models and enable them to adapt to each individual in a personalized manner, without the need to wait for all the test individuals to arrive.
Moreover, SF-UIDA also adopts a source-free UDA strategy, which is a more practical setup that does not require accessing the source data, thereby lowering the time cost and protecting the data privacy. As shown in Fig. \ref{fig:0} (b), our framework enables the source model to rapidly adapt to each new target in a personalized and plug-and-play fashion without accessing the source data.

Our SF-UIDA is evaluated on three public datasets. The process of customization is efficient in time and acceptable in clinics, only taking a short amount of time to transform the source model into a personalized model. Besides, it is worth pointing that our SF-UIDA framework can be easily implemented without any modification to the source model structure, achieving plug-and-play application in practice. Our contributions are as follows:
\begin{itemize}
	\item 
	We devise a novel Source-Free Unsupervised Individual Domain Adaptation framework for automatic sleep staging, named SF-UIDA. It meets the practical needs for a plug-and-play personalized customization that can be applied to each newly appeared individual without accessing the source data.
	\item 
	We propose a two-step subject-specific alignment strategy to mitigate the impact of individual discrepancies. It effectively transforms the source model into a personalized model for each new individual within a short adaptation time. 
	\item
	Our SF-UIDA achieves the best generalization performance across three public sleep staging datasets, compared to other methods.
\end{itemize}
\section{RELATED WORK}
\subsection{Automatic Sleep Staging}
There have been many models proposed for automatic sleep staging in recent years. Some studies commonly employ convolutional networks to extract local sleep features. For example, U-time \cite{utime, usleep} is a fully CNN network based on the U-net architecture that can excellently model sleep-related features. SalientSleepNet \cite{jia2021salientsleepnet} is also a fully CNN network based on U$^2$-net which can capture multi modal sleep feature. Considering the advantage of capturing long-term temporal information, there are also some studies utilizing recurrent neural networks \cite{mikolov2010recurrent} or transformer encoders \cite{vaswani2017attention} for sleep staging. \citet{Phan_2022} proposed SleepTransformer, a transformer-based sequence-to-sequence model that improves the interpretability of the sleep-staging task. Although these models obtain good performance for sleep staging, they have not taken individual discrepancies into account, leading to dramatically degraded performance when applied to target domains (i.e., unknown subjects). In this paper, we propose a new training strategy to address individual discrepancies for sleep staging, so as to improve the model generalization ability in practice.
\subsection{Source-free Unsupervised Domain Adaptation}
To address the challenge of model generalization on unseen data, some studies have employed UDA methods to facilitate knowledge transfer between the source and target domains. Existing UDA studies \cite{tang2022deep, wang2024generalizable} have reduced domain shift and extracted domain-invariant features through distance-based alignment. \citet{fan2022unsupervised} utilized statistical alignment to mitigate domain shift across several sleep datasets. These UDA methods effectively enhance the generalization ability of the source model for sleep staging. However, they rely on source data for joint training, which is impractical and time-consuming for each new subject in real-life scenarios. In contrast, source-free UDA \cite{liang2020we} offers a more practical solution by eliminating the need for source data during adaptation. Contrastive learning (CL) \cite{cpc, simclr, chen2021exploring} is commonly used in source-free UDA, focusing on mining intrinsic representation features within the data. \citet{chen2022contrastive} applied a self-supervised CL method to facilitate target feature learning and achieve test-time adaptation, while \citet{wang2022continual} utilized weight-averaged and augmentation-averaged predictions to generate pseudo-labels for adaptation. These source-free approaches enable rapid adaptation of the source model to target domains. However, they often overlook individual discrepancies and treat the entire target domain, encompassing multiple subjects, as a single distribution for adaptation. This can hinder accurate predictions for individuals whose distributions significantly deviate from the overall distribution. To address this, we propose the SF-UIDA framework, which accounts for individual discrepancies and enables personalized customization for each individual.
\section{METHODOLOGY}
\begin{figure}[!tb]
	\centering
	\includegraphics[width=1.0\columnwidth]{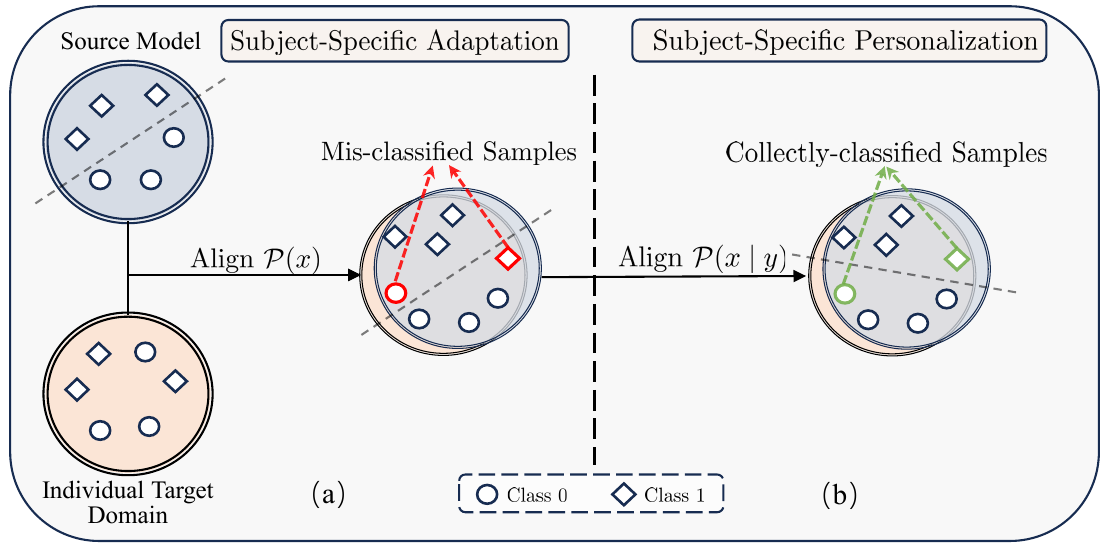}
	\caption{Illustration of the two-step alignment strategy in the SF-UIDA Framework: aligning the marginal probability distribution 
	$\mathcal{P}_\mathcal{T}(x)$ and the class conditional probability distribution $\mathcal{P}_\mathcal{T}(x\mid{y})$
		for individual target domains}
	\label{fig:8}
\end{figure}
\subsection{Problem Formulation}
In this work, we try to address the issue of individual discrepancy for the task of automatic sleep staging by employing an UDA-based approach. Here, we introduce the concept of individual target domain, which consists of the recordings from only one subject. Formally, given a labeled source domain $\mathcal{D}_\mathcal{S}$=${\{\mathcal{X}_\mathcal{S}^i, \mathcal{Y}_\mathcal{S}^i\}}_{i=1}^{\mathcal{N}_\mathcal{S}}$ with $\mathcal{N}_\mathcal{S}$ subjects and an unlabeled individual target domain $\mathcal{D}_\mathcal{T}$=${\{\mathcal{X}_\mathcal{T}^j\}}_{j=1}^{\mathcal{N}_\mathcal{T}}$ with ${\mathcal{N}_\mathcal{T}} = 1$. We denote the distributions of different domains as $\mathcal{P}_\mathcal{S}(x,y)$ and $\mathcal{P}_\mathcal{T}(x,y)$ respectively, where $\mathcal{P}_\mathcal{S}(x,y) \neq \mathcal{P}_\mathcal{T}(x,y)$. We employ the sleep sequence $\mathcal{X}_\mathcal{S}$=$(x_1^\mathcal{S},x_2^\mathcal{S},x_3^\mathcal{S},...,x_L^\mathcal{S})$ of length $L$ and its corresponding label $\mathcal{Y}_\mathcal{S}$=$(y_1^\mathcal{S},y_2^\mathcal{S},y_3^\mathcal{S},...,y_L^\mathcal{S})$ from $\mathcal{D}_\mathcal{S}$ as inputs, where $x_i^\mathcal{S}$, $y_i^\mathcal{S}$ denotes the data and label of $i$-$th$ epoch in the sequence $\mathcal{X}_\mathcal{S}$. Our main purpose is to accurately predict the label $\mathcal{Y}_\mathcal{T}$=$(y_1^\mathcal{T},y_2^\mathcal{T},y_3^\mathcal{T},...,y_L^\mathcal{T})$ of unlabeled individual target domain sequential sample  $\mathcal{X}_\mathcal{T}$=$(x_1^\mathcal{T},x_2^\mathcal{T},x_3^\mathcal{T},...,x_L^\mathcal{T})$.
\subsection{Overview}
In this work, we first pretrain the source model from a labeled source domain $\mathcal{D}_\mathcal{S}$. The probability distribution of the source domain can be described as follows.
\begin{equation}\label{prob}
	\mathcal{P}_\mathcal{S}(x,y) = \mathcal{P}_\mathcal{S}(x)\mathcal{P}_\mathcal{S}(y\mid{x})= \mathcal{P}_\mathcal{S}(y)\mathcal{P}_\mathcal{S}(x\mid{y})
\end{equation}
Considering the individual discrepancies among each subject, our SF-UIDA framework contains a two-step alignment: \textbf{subject-specific adaptation} and \textbf{subject-specific personalization}, to align each of the individual target distribution $\mathcal{P}_\mathcal{T}$ with the source domain distribution $\mathcal{P}_\mathcal{S}$, customizing the sleep staging model for each individual. Specifically, we use the unlabeled data from each individual target domain and the generated pseudo-labels to make the source model's probability distribution $\mathcal{P}_\mathcal{S}(x,y)$ align with the individual target domain's probability distribution $\mathcal{P}_\mathcal{T}(x)$ and class-conditional probability distribution $\mathcal{P}_\mathcal{T}(x\mid{y})$ illustrated in Fig. \ref{fig:8}.
The whole process can be totally divided into three training stages:
$\textbf{Source Model Pretraining}$: We employ several classical sleep stage classification (SSC) models as the pretraining models to learn the general sleep features from the source domain. %Each SSC model consists of two parts: a feature extractor and a temporal encoder, capable of extracting sleep features and encoding temporal information.
$\textbf{Subject-Specific Adaptation}$: We propose a subject-specific adaptation by proposing a sequential cross-view prediction task on individual target domain. It is used to capture subject-specific sleep representations and align with the individual target domain's probability distribution $\mathcal{P}_\mathcal{T}(x)$, mitigating the impact of individual discrepancies on the pretrained source model.
$\textbf{Subject-Specific Personalization}$: We employ a teacher model based pseudo-labeling strategy for fine-tuning, so as to learn the fine-grained distribution of different classes in individual target domains. It enables the source model further align the class-conditional probability distribution $\mathcal{P}_\mathcal{T}(x\mid{y})$, thus achieving personalized customization. \textbf{Notably, our model customization is absolutely source-free, using only the unlabeled target data for personalization.}
\subsection{Source Model Pretraining}
%We first extract general characteristics from PSG recordings by conducting pretraining model on the source domain. Specifically, we employ three classical lightweight sleep staging models as the pretrained models. Each model is equipped with a specifically designed feature extractor to extract sleep features and a temporal encoder to learn the temporal information from the sleep sequences. The source model parameters will be transferred to the subsequent two-step alignment process.
We first extract general characteristics from PSG recordings by pretraining a model on the source domain. Specifically, we employ three classical lightweight sleep staging models as pretrained models. Each model has a specialized extractor for sleep features and a temporal encoder to capture temporal information from sleep sequences. The parameters of the source model are then transferred to the subsequent  process.
\subsection{Subject-Specific Adaptation}
%Due to the independence of the individuals from test set and the existence of inter-individual differences, each individual target domain exhibits an independent distribution. Its marginal distribution may significantly deviate from the source domain distribution. Therefore, the pretrained source model cannot generalize well to individual target domains. \textbf{In this step, our objective is to reduce the domain shift and mitigate the impact of individual discrepancies without access to the source data.} Motivated by the CPC, TS-TCC \cite{cpc, tstcc} algorithms, we propose a subject-specific adaptation scheme to perform unsupervised domain adaptation separately for each individual target domain, aligning the probability distribution $\mathcal{P}_\mathcal{T}(x)$ by conducting a complex sequential cross-view prediction task. By this way, we can model the specific representations for each individual.
Due to the independence of individuals in the test set and inter-individual differences, each individual target domain exhibits a distinct distribution, often significantly deviating from the source domain distribution. Therefore, the pretrained source model struggles to generalize to individual target domains. \textbf{In this step, our objective is to reduce domain shift and mitigate the impact of individual discrepancies without access to the source data.} Inspired by the CPC and TS-TCC \cite{cpc, tstcc} algorithms, we propose a subject-specific adaptation scheme for unsupervised domain adaptation, aligning the probability distribution $\mathcal{P}_\mathcal{T}(x)$ through a complex sequential cross-view prediction task. This approach enables us to model specific representations for each individual.

\subsubsection{Sequential Cross-View Contrasting}
\begin{figure}[tb]
	\centering
	\includegraphics[width=1.0\columnwidth]{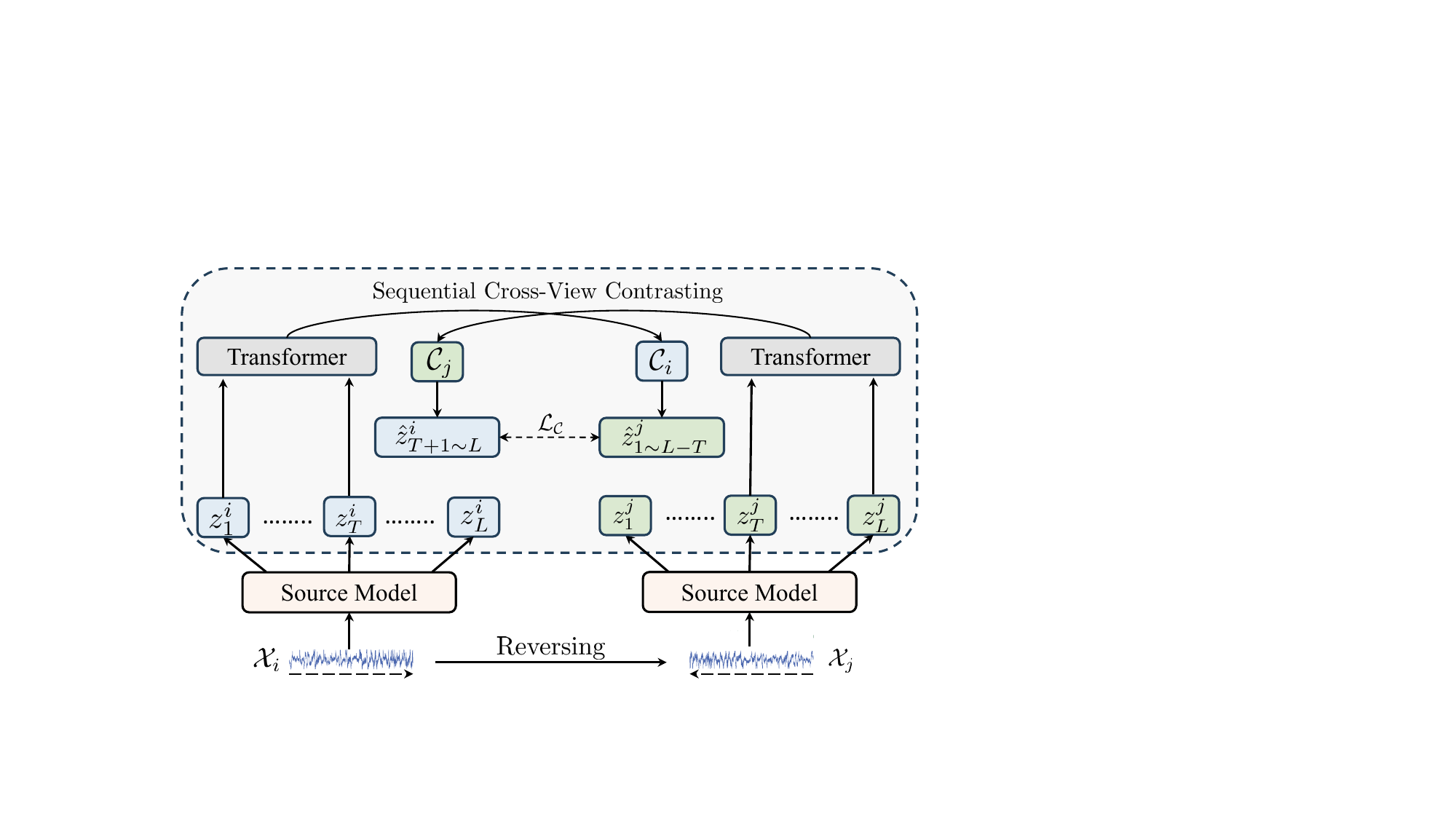}
	\caption{The architecture of the proposed SCC module.}
	\label{fig:3}
\end{figure}

According to the AASM\cite{aasm}, the transition patterns of sleep stages between neighboring epochs are crucial for accurate sleep staging. \textbf{These transitions occur not only in the forward direction (e.g., W→N1→N2→N3) but also in reverse (e.g., REM←N3←N2).} Motivated by these patterns, we propose a novel \textbf{S}equential \textbf{C}ross-view \textbf{C}ontrasting (\textbf{SCC}) module to model the bidirectional transition relationships within subject-specific sequences, as illustrated in Fig. \ref{fig:3}.
To generate a new augmented view, reversing the original sequence effectively contrasts the temporal relationships between different views. Formally, given an input sleep sequence $\mathcal{X}_i$=$(x_1,x_2,...,x_{L-1}, x_L)$, its augmented view $\mathcal{X}_j$=$(x_L,x_{L-1},..., x_2, x_1)$ is obtained by reversal. After feeding $\mathcal{X}_i$ and $\mathcal{X}_j$ into the Feature Extractor and Feature Encoder, we obtain their corresponding latent representations $\mathcal{Z}_i$=$(z_1^i,z_2^i,...,z_{L-1}^i, z_L^i)$ and $\mathcal{Z}_j$=$(z_L^j,z_{L-1}^j,...,z_2^j, z_1^j)$, respectively. 
For a given time step $T$ $(1<T<L)$, we utilize a transformer as an autoregressive model to encode ${\mathcal{Z}_{i:}}_{{t}\leq{T}}$, ${\mathcal{Z}_{j:}}_{{L-T}\leq{t}\leq{L}}$ into a contextual vector: $\mathcal{C}_i$ and $\mathcal{C}_j$. We then establish a sequential cross-view task, using linear layers to predict the future $L-T$ sleep timesteps from $z_{T+1}^i$ to $z_L^i$ in sequence $\mathcal{Z}_i$ by leveraging contextual vector $\mathcal{C}_j$, such that $\hat{z}_{T+k}^i = f_{T+k}(\mathcal{C}_j)$, where $\hat{z}_{T+k}^i$ denotes the predicted timesteps for $z_{T+k}^i$ and $f_{T+k}$ is the corresponding predicting linear layer. Similarly, we use contextual vector $\mathcal{C}_i$ to predict the past $L-T$ sleep timesteps from $z_1^j$ to $z_{L-T}^j$ in sequence $\mathcal{Z}_j$. 
The cross-view predicted timesteps can be formulated as $\hat{\mathcal{Z}}_{T+1 \sim L}^i = (f_{T+1}(\mathcal{C}_j), f_{T+2}(\mathcal{C}_j),..., f_{L}(\mathcal{C}_j) ) = (\hat{z}_{T+1}^i, \hat{z}_{T+2}^i,..., \hat{z}_{L}^i)$ and $\hat{\mathcal{Z}}_{1 \sim L-T}^j = (f_{1}(\mathcal{C}_i), f_{2}(\mathcal{C}_i),..., f_{L-T}(\mathcal{C}_i)) = (\hat{z}_{1}^j, \hat{z}_{2}^j,..., \hat{z}_{L-T}^j)$, respectively.
We then apply the Maximum Mean Discrepancy (MMD) loss to minimize the distance between the cross-view predicted timesteps $\hat{\mathcal{Z}}_i$ and $\hat{\mathcal{Z}}_j$ as follows:

\begin{equation}\label{Lci}
	\mathcal{L}_{\mathcal{C}} = -\frac{1}{\mathcal{K}}\sum_{k=1}^{\mathcal{K}}{\mathcal{D}_{\mathcal{MMD}}(\hat{z}_{T+k}^i, \hat{z}_{k}^j)}
\end{equation}
where $\mathcal{K}$ denotes the $L-T$. Notably, for a specific predicted timestep $\hat{z}_{k}^j$, we minimize its MMD distance between the corresponding predicted timestep $\hat{z}_{T+k}^i$ rather than $\hat{z}_{k}^i$. This is because after reversing, the position corresponding to $z_{T+k}^i$ aligns perfectly with $z_{k}^j$. 
\begin{algorithm}[tb]
	\DontPrintSemicolon
	\caption{Source-Free Unsupervised Individual Domain Adaptation algorithm}
	\KwIn{$\mathcal{X}_\mathcal{S}$, $\mathcal{Y}_\mathcal{S}$, $\mathcal{X}_\mathcal{T}$}
	\KwOut{$\mathcal{F}_\theta$}
	\textbf{Source Model Pretraining:}\par
	Pretrain the source model by $\mathcal{X}_\mathcal{S}$, $\mathcal{Y}_\mathcal{S}$ and return $\mathcal{F}_\theta$. \par
	\textbf{Subject-Specific Adaptation:}\par
	Generate the augmented view $\mathcal{X}_j$ for each input target sleep sequence $\mathcal{X}_i$.\par
	\For{$i=1$ to $n$}{
		Compute latent representations $\mathcal{Z}_i$, $\mathcal{Z}_j$. \par
		Compute contextual vectors $\mathcal{C}_i$, $\mathcal{C}_j$. \par
		Optimize $\mathcal{F}_\theta$ by minimizing Eq.\ref{Lci}. \par}
	return retrained model $\mathcal{F}_\theta$. \par
	\textbf{Subject-Specific Personalization:}\par
	Initialize the teacher model $\mathcal{F}_{\theta_T}$ and transfer the weights $W_\theta$ to $W_{\theta_T}$. \par
	\For{$i=1$ to $n$}{
		Generate confident pseudo sequence labels $\hat{\mathcal{Y}_\mathcal{T}}$ by $\mathcal{F}_{\theta_T}$ by Eq. \ref{teacher_y}, Eq. \ref{teacher_label}.\par
		Optimize $\mathcal{F}_\theta$ by minimizing Eq.\ref{cross_entropy}.\par
		Update $W_{\theta_T}$ by Eq.\ref{ema}
	}
	return $\mathcal{F}_\theta$.
\end{algorithm}
\subsection{Subject-Specific Personalization}
The subject-specific adaptation aligns the marginal distributions between the source domain and the individual target domain. Due to the influence of individual differences, there still exist class-conditional distribution discrepancies between each individual target domain and the source domain, which may lead to erroneous alignment of different classes across the source and target domains shown in Fig.\ref{fig:8} (a). Conventional solutions typically rely on fine-tuning with labeled data to address this issue \cite{eldele2023self}. However, we are unable to employ supervised methods for class-conditional distribution alignment as the target labels are unavailable. Inspired by \citet{teacher} and \citet{ragab2022self}, we employ a teacher model based on pseudo-label generation approach to tackle this problem shown in Fig.\ref{fig:8}(b). Notably, we have introduced \textbf{sequence confidence} for each individual, producing robust pseudo sequence labels. We solely preserve the confident ones for further fine-tuning, which enable us to better align the class-conditional distribution with the source domain and model the personalized representations. Specifically, we migrate the model parameters $\mathcal{W}_{\theta}$ to the teacher model $\mathcal{F}_{\theta_T}$ using Exponential Moving Average (EMA). The updates to the teacher model parameters are as follows:
\begin{table*}[tb]
	\centering 
	\setlength{\tabcolsep}{2mm}
		\begin{tabular}{lccccccl}
			\toprule[1.5pt]
			Dataset      & Subjects & Subjects   & CV & Sampling & Scoring & EOG Channels & EEG Channels    \\
				         & (all) & (we choose)   & (fold) & (Hz) & (standard) & (we choose) & (we choose)  \\
			 \midrule[1pt]
			
			ISRUC        & 100      & 98  & 10 & 200                             & AASM     & E1, E2 & F3, F4, C3, C4, O1, O2\\
			HMC          & 151     & 145 & 10 & 256                             & AASM     & E1, E2 & F4, C4, O2, C3\\
			SleepEDF-153 & 78     & 78 & 10 & 100                             & R\&K     & EOG horizontal & Fpz-Cz, Pz-Oz\\ \bottomrule[1.5pt] 
	\end{tabular}
	\caption{A brief description about three public sleep staging datasets.}
	\label{table:1}
\end{table*}
\begin{equation}\label{ema}
	\mathcal{W}_{\theta_T} = \alpha \mathcal{W}_{\theta_T} + (1-\alpha)\mathcal{W}_{\theta}
\end{equation}
where $\mathcal{W}_{\theta_T}$ denotes the parameters of the teacher model $\mathcal{F}_{\theta_T}$ and $\alpha$ is a hyper-parameter employed to regulate the update rate of the teacher model parameters. For each sleep sequence $\mathcal{X}_\mathcal{T}$=$(x_1,x_2,x_3,...,x_L)$, we can obtain the corresponding predicted sequence probabilities $\mathcal{Y}_\mathcal{T}$=$(y_1,y_2,y_3,...,y_L)$ by the teacher model. 
Given a sleep sequence $\mathcal{X}_\mathcal{T}$ with the length $L$, we retain it for subsequent fine-tuning iff there are $\mathcal{N}_c$ or more epochs in the sequence and the prediction probabilities of each epoch not less than the confidence threshold $\xi$. It can be formalized as follows:
\begin{equation}\label{teacher_y}
	\mathcal{Y}_\mathcal{T} = softmax(\mathcal{F}_\theta(\mathcal{X}_\mathcal{T}))
\end{equation}
\begin{equation}\label{teacher_label}
	\hat{\mathcal{Y}_\mathcal{T}} = \mathds{1}_{s}{ ([\sum_{i=1}^L{\mathds{1}_{e}{(max(y_i^\mathcal{T})>{\xi})}} ]\geq \mathcal{N}_c)}\cdot{\mathcal{Y}_\mathcal{T}}
\end{equation}
where $\mathds{1}_{e}$ is the confident epoch indicator function, evaluating to 1 iff $max(y_t^k)>{\xi}$ and $\mathds{1}_{s}$ is the confident sequence indicator function, evaluating to 1 iff there are $\mathcal{N}_c$ or more confident epochs $y_i^\mathcal{T}$ in the $\mathcal{Y}_\mathcal{T}$. In most existing studies, the confidence threshold $\xi$ is set to be greater than 0.9. However, based on our confident sequence setting, we focus more on the overall confidence of the sequence rather than that of a single epoch. So we need a higher tolerance for the threshold $\xi$ and we set the it equal to 0.8 and the $\mathcal{N}_c$ is set to 15. 
To ensure alignment of the class-conditional distribution, we employ the confident sequence labels to finetune the model, by using a cross-entropy loss:
\begin{equation}\label{cross_entropy}
	\mathcal{L}_{ce} = -\mathds{E}_{\mathcal{X}_\mathcal{T}\sim{\mathcal{P}_\mathcal{T}}}[\sum_{k=1}^K{\mathcal{Y}_\mathcal{T}^klog(\hat{\mathcal{Y}_\mathcal{T}^k})}]
\end{equation}
	During the whole process, only the unlabeled target individual's data is needed for personalized customization without access to the source data. The algorithm of SF-UIDA framework is illustrated in Algorithm 1. 
\section{EXPERIMENTS}
\subsection{Datasets}
%As shown in Tab.\ref{table:1}, we evaluated our approach on three publicly available datasets, namely Sleep-EDF \cite{sleepedf}, ISRUC \cite{isruc}, HMC \cite{hmc}. For each dataset, we utilize both EEG and EOG channels as input. \textbf{ISRUC}   A public database consists of three sub-groups. We specifically selected sub-group 1, which consists of all-night polysomnography (PSG) recordings from 100 adult individuals and contains 86400 samples. We excluded subject 8 and 40 due to some missing channels. \textbf{SleepEDF-153} A public Physionet database consists of 78 healthy subjects aged 25-101 and contains 188,760 samples. All the subjects' recordings were used for evaluation.\textbf{HMC}  A public dataset includes a total of 151 subjects from the sleep center database of the Haaglanden Medisch Centrum (The Netherlands). It consists of 129440 samples. We excluded subjects 14,32,33,64,112,135 due to some missing channels. All the sleep recordings used in our experiments were bandpass filtered (0.3Hz–35Hz) and resampled to 100Hz.
As shown in Table \ref{table:1}, we evaluated our approach on three publicly available datasets: Sleep-EDF \cite{sleepedf}, ISRUC \cite{isruc}, and HMC \cite{hmc}. For each dataset, we utilized both EEG and EOG channels as input.
\textbf{ISRUC}: This public database consists of three sub-groups. We specifically selected sub-group 1, which includes all-night polysomnography (PSG) recordings from 100 adults, totaling 86,400 samples. Subjects 8 and 40 were excluded due to missing channels.
\textbf{SleepEDF-153}: A public Physionet database comprising 78 healthy subjects aged 25 to 101, containing 188,760 samples. All subjects' recordings were used for evaluation.
\textbf{HMC}: This public dataset includes recordings from 151 subjects at the Haaglanden Medisch Centrum (The Netherlands), consisting of 129,440 samples. Subjects 14, 32, 33, 64, 112, and 135 were excluded due to missing channels. All sleep recordings were bandpass filtered (0.3 Hz–35 Hz) and resampled to 100 Hz.
\begin{table*}[tb]
	\centering 
	 \setlength{\tabcolsep}{0.8mm}
	 \footnotesize
	\begin{tabular}{ccccccccccccccccccccl}
		\toprule[1.5pt]
		\multicolumn{1}{l}{} & \multicolumn{6}{c}{ISRUC}                                                        & \multicolumn{6}{c}{HMC}                                                                               & \multicolumn{6}{c}{SleepEDF}                                                                         & \multicolumn{2}{c}{\multirow{2}{*}{Average}} \\ \cmidrule[1pt](lr){2-7} \cmidrule[1pt](lr){8-13} \cmidrule[1pt](lr){14-19}
		& \multicolumn{2}{c}{Deep.} & \multicolumn{2}{c}{Tiny.} & \multicolumn{2}{c}{Rec.} & \multicolumn{2}{c}{Deep.} & \multicolumn{2}{c}{Tiny.} & \multicolumn{2}{c}{Rec.} & \multicolumn{2}{c}{Deep.} & \multicolumn{2}{c}{Tiny.} & \multicolumn{2}{c}{Rec.} & \multicolumn{2}{c}{}                         \\ \cmidrule[1pt](lr){2-21}
		& ACC         & MF1         & ACC         & MF1         & ACC         & MF1        & ACC             & MF1            & ACC             & MF1            & ACC            & MF1            & ACC             & MF1            & ACC             & MF1            & ACC            & MF1            & ACC         & \multicolumn{1}{c}{MF1}        \\ \midrule[1pt]
		Source Only          & 59.7        & 49.7        & 66.9        & 57.2        & 64.1        & 57.3       & 71.8            & 66.0             & 72.2            & 64.5           & 74.0             & 66.8           & 51.8            & 35.5           & 61.4            & 46.7           & 73.6           & 61.2           & 66.2        & 56.1                           \\
		CPC                  & 67.3        & 62.1        & 69.3        & 62.0          & 69.0          & 63.1       & 78.1            & 74.0             & 74.9            & 68.3           & 76.4           & 70.2           & 78.5            & 69.2           & 77.8            & 65.2           & 78.0             & 66.1           & 74.4        & 66.7                           \\
		SimSiam              & 68.4        & 63.5        & 69.2        & 62.6        & 69.4        & 62.8       & 78.2            & 73.8           & 75.7            & 69.4           & 76.6           & 70.9           & 80.6            & 71.0             & 77.9            & 64.8           & 78.1           & 66.2           & 74.9        & 67.2                           \\
		Adacontrast          & 68.6        & 63.7        & 69.7        & 63.2        & 69.6        & 62.9       & 78.1            & 73.7           & 75.4            & 69.2           & 76.8           & 71.1           & 80.5            & 70.9           & 77.6            & 64.5           & 77.9           & 65.7           & 74.9        & 67.2                           \\
		CoTTA                & 67.8        & 62.8        & 70.2        & 62.9        & 69.3        & 62.8       & 77.7            & 73.5           & 75.0              & 68.3           & 76.4           & 70.2           & 80.2            & 70.7           & 77.9            & 65.4           & 78.0             & 66.1           & 74.7        & 67.0                           \\
		C-SFDA               & 68.6        & 63.7        & 70.0          & 63.5        & 70.4        & 63.4       & 78.1            & 73.7           & 75.5            & 69.4           & 76.8           & 71.1           & 80.3            & 70.8           & 77.6            & 64.4           & 77.9           & 65.6           &  75.0        & 67.3                           \\ \midrule
		\textbf{Ours}                & \textbf{70.1}            & \textbf{64.7}           & \textbf{72.2}            & \textbf{65.1}           & \textbf{71.5}           & \textbf{64.6}           & \textbf{79.1}            & \textbf{75.0}           & \textbf{76.6}            & \textbf{70.4}           & \textbf{77.8}          & \textbf{72.3}           & \textbf{81.6}           & \textbf{72.2}           & \textbf{79.3}            & \textbf{66.5}           & \textbf{79.1}           & \textbf{67.4}           & \textbf{76.4}        & \textbf{68.7}         \\ \bottomrule[1.5pt]                 
	\end{tabular}
	\caption{Performance comparison with existing source-free UDA methods. Notably, Deep., Tiny., and Rec. refer to DeepSleepNet, TinySleepNet, and RecSleepNet, respectively.}
	\label{tab:4}
\end{table*}
\subsection{Settings}
\subsubsection{Baseline Models}
We need to select baseline models to evaluate our SF-UIDA framework. Considering the efficiency requirement in clinical practice, the fine-tuning for each individual cannot be time-consuming. Therefore, \textbf{we selected three lightweight sleep staging models from existing studies}, each of which is comprised of the feature extractor and the temporal encoder:
$\textbf{DeepSleepNet}$ \cite{Supratak_2017}: a classical CNN-BiLSTM model for extracting sleep features and learning transition rules.
$\textbf{TinySleepNet}$ \cite{tinysleepnet}: a more lightweight model based on the DeepSleepNet.
$\textbf{RecSleepNet}$ \cite{recsleepnet}: a CNN-LSTM model based on feature representation reconstruction. 
Here, we do not choose some other sleep staging models, which are also classical and perform well, such as UTime \cite{utime}, SalientSleepNet \cite{jia2021salientsleepnet} or SleepTransformer \cite{Phan_2022}, because their network structures are complex and not suitable for our problem in clinical practice.
\subsubsection{Implementation}
Based on the publicly available source code, we re-implemented the three baseline models using pytorch.
The experimental settings are as follows:
\textbf{Source Model Pretraining}: The pretraining epoch is set to 100. The learning rate is set to 1e-4.
\textbf{Subject-Specific Adaptation}: The training epoch of this stage is set to 5. The learning rate is set to 1e-7. The time step $T$ is set to 17.
\textbf{Subject-Specific Personalization}: The fine-tuning epoch is set to 10. The learning rate is set to 1e-7. The momentum $\alpha$ is set to 0.996. 
We use the Adam optimizer to train the model, the $\beta$ is set to [0.5,0.99], the weight decay is set to 3e-4, the size of mini-batch is set to 32. 
The model is trained on a single machine equipped with an Intel Core i9 10900K CPU and eight NVIDIA RTX 3080 GPUs. The source code is publicly available\footnote{https://github.com/xiaobaben/SF-UIDA}.
\subsubsection{Performance Measurement}
We employ 10-fold cross-validation (CV) to evaluate the performance of our approach across three different datasets. Different from the conventional settings in previous studies, which only included training and validation sets, \textbf{we divided the dataset into the training, validation, and test sets and the ratio is 8:1:1}. The test set is regarded as unknown subjects, \textbf{where there are no repetitive individuals in the test set, ensuring that each individual appears only once in the test set throughout the 10-fold CV experiment}. In each fold, we employ the training and validation sets to pretrain the source model. Subsequently, \textbf{the source model goes though personalization customization on each individual in the test set.} Finally, we compute the average metrics for each individual in the test set. We employ Accuracy (ACC) and Macro-F1 score (MF1) as evaluation metrics.
\subsection{Result Analysis}
\subsubsection{Compared with Other Existing Methods}
We compare our method with other classical source-free UDA methods in sleep staging, to further investigate the generalization ability of SF-UIDA.
$\textbf{Source only}$ : A method to directly test the individual target domain using source model.
$\textbf{CPC}$ \cite{cpc}: A classical contrastive learning approach to learn representation of time-seires data by predicting the future timesteps. $\textbf{SimSiam}$ \cite{chen2021exploring}: An efficient contrastive learning approach, which focuses on representation learning using stop-gradient strategy and symmetrized loss.  
$\textbf{Adacontrast}$ \cite{chen2022contrastive}: A test-time adaptation method using contrastive learning to facilitate target feature learning. $\textbf{CoTTA}$ \cite{wang2022continual}: A test-time adaptation method  , which can effectively adapt off-the-shelf source pretrained models to target domains. $\textbf{C-SFDA}$ \cite{karim2023c}: A curriculum learning aided self-training framework for SFDA is designed to adapt efficiently and reliably to target domains.
We implement these SFDA methods within our framework, and the performance comparison is shown in Table \ref{tab:4}. Our method outperforms existing approaches, underscoring the effectiveness of our personalized customization strategy in enhancing overall performance. Compared to the Source Only method, our approach significantly enables the source model to adapt to individual target domains, achieving model customization and improved performance.
%We implement these selected SFDA methods based on our settings, and the performance comparison is shown in Tab.\ref{tab:4}. Our method demonstrates superior performance compared to existing methods, highlighting the effectiveness of our proposed personalized customization approach in enhancing overall performance. When compared to source only method, our method can significantly enable the source model adapt to the individual target domains, achieving model customization and improving the overall performances. 
Among the compared methods, it is worth noting that methods leveraging contrastive learning(e.g. SimSiam, Adacontrast, C-SFDA) exhibit better performance compared to other approaches (e.g. CPC, CoTTA). 
\begin{table*}[tb]
	\centering 
	 \setlength{\tabcolsep}{0.8mm}
	\footnotesize
		\begin{tabular}{lccccccccccccccccccll}
			\toprule[1.5pt]
			& \multicolumn{6}{c}{ISRUC}                                                                             & \multicolumn{6}{c}{HMC}                                                                                                         & \multicolumn{6}{c}{SleepEDF}                                                                                                                                                      & \multicolumn{2}{c}{}                              \\ \cmidrule[1pt](lr){2-7} \cmidrule[1pt](lr){8-13} \cmidrule[1pt](lr){14-19}
			\multicolumn{1}{c}{} & \multicolumn{2}{c}{Deep.} & \multicolumn{2}{c}{Tiny.} & \multicolumn{2}{c}{Rec.} & \multicolumn{2}{c}{Deep.} & \multicolumn{2}{c}{Tiny.} & \multicolumn{2}{c}{Rec.}                           & \multicolumn{2}{c}{Deep.}                          & \multicolumn{2}{c}{Tiny.}                          & \multicolumn{2}{c}{Rec.}                           & \multicolumn{2}{c}{\multirow{-2}{*}{Average}}     \\ \cmidrule[1pt](lr){2-21}
			\multicolumn{1}{c}{} & ACC             & MF1            & ACC             & MF1            & ACC            & MF1            & ACC             & MF1            & ACC             & MF1            & ACC                         & MF1                         & ACC                         & MF1                         & ACC                         & MF1                         & ACC                         & MF1                         & \multicolumn{1}{c}{ACC} & \multicolumn{1}{c}{MF1} \\ \midrule[1pt]
			SO                   & 59.7            & 49.7           & 66.9            & 57.2           & 64.1           & 57.3           & 71.8            & 66.0           & 72.2            & 64.5           & 74.0                        & 66.8                        & 51.8                        & 35.5                        & 61.4                        & 46.7                        & 73.6                        & 61.2                        & 66.2                    & 56.1                    \\
			SO+SSA               & 69.0            & 63.8           & 71.3            & 63.4           & 70.3           & 63.5           & 77.5            & 73.1           & 75.5            & 69.0           & 76.6                        & 70.7                        & 79.4                        & 69.8                        & 76.1                        & 62.8                        & 77.2                        & 64.9                        & 74.8                    & 66.8                    \\
			SO+SSP               & 68.0            & 62.4           & 70.9            & 64.7           & 68.0           & 62.4           & 78.4            & 74.3           & 74.5            & 70.3           &  77.7 & 71.9 & 79.5 &  69.9 &  78.9 & 66.4 & 79.1 & 67.4 & 75.0                    & 67.7                    \\
			SO+SSA+SSP                        & \textbf{70.1}            & \textbf{64.7}           & \textbf{72.2}            & \textbf{65.1}           & \textbf{71.5}           & \textbf{64.6}           & \textbf{79.1}            & \textbf{75.0}           & \textbf{76.6}            & \textbf{70.4}           & \textbf{77.8}          & \textbf{72.3}           & \textbf{81.6}           & \textbf{72.2}           & \textbf{79.3}            & \textbf{66.5}           & \textbf{79.1}           & \textbf{67.4}           & \textbf{76.4}        & \textbf{68.7} \\ \bottomrule[1.5pt]             
	\end{tabular}
	\caption{Ablation experiment overview.}
	\label{table:ablation}
\end{table*}

\subsubsection{Compared with Non-Personalized Domain Adaptation}
\begin{figure}[!tb]
	\centering
	\includegraphics[width=1.0\columnwidth]{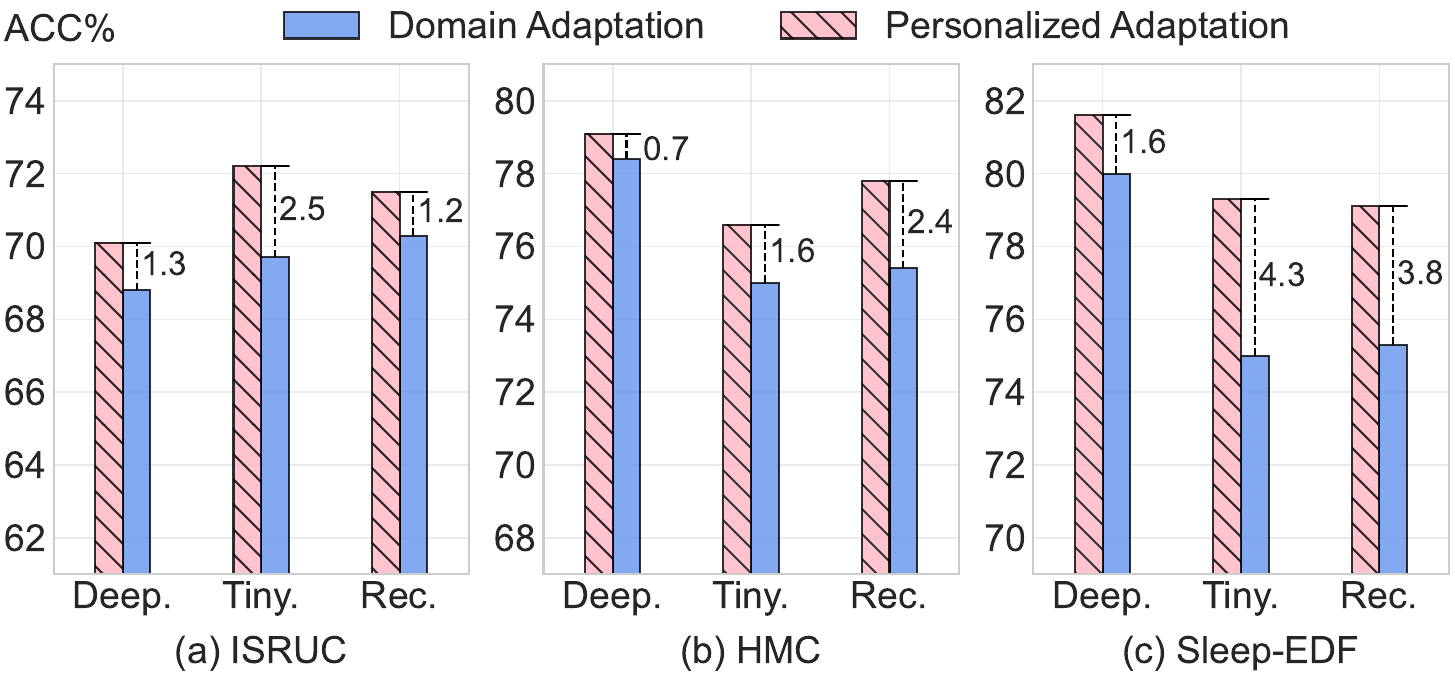}
	\caption{Comparison with traditional domain adaptation paradigm (i.e., non-personalized domain adaptation).}
	\label{fig:4}
\end{figure}

The traditional domain adaptation (DA) paradigm considers a batch of subjects as the target domain for subsequent adaptation which is impractical in real life. In contrast, our individual domain adaptation based method allows for plug-and-play application on each new subject without waiting. To evaluate the effectiveness of our individual DA setting compared to traditional DA paradigm, we conducted a comparative study. Notably, we maintained the consistent partitioning of the 10-fold cross validation and the SF-UIDA framework. The only difference is: \textbf{for the traditional DA paradigm, in each fold we use the data of all target individuals (i.e., test set) to perform adaptation, rather than conducting model customization for each individual separately.} We evaluated the performance of the fine-tuned model on each test individual shown in Fig. \ref{fig:4}. Compared to the traditional domain adaptation paradigm, our personalized adaptation based paradigm can achieve comprehensive superiority in performance on all baseline models across the three datasets. It proves that our method can not only meet the practical needs in real-world scenarios, where the personalized customization can be plug-and-play applied to new individuals, but also better achieve individual performance improvement after adaptation.
\subsubsection{Ablation Study}
To investigate the importance of our proposed two-step alignment strategy, we conducted this ablation experiments. The model variants are defined as follows:\par
\begin{itemize}
	\item 
	\textbf{SO}: which means source only that we directly use the source model for testing.\par
	\item 
	\textbf{SO+SSA}: only the \textbf{S}ubject-\textbf{S}pecific \textbf{A}daptation (SSA) stage is preserved in the SF-UIDA framework.\par
	\item 
	\textbf{SO+SSP}: only the \textbf{S}ubject-\textbf{S}pecific \textbf{P}ersonalization (SSP) stage is preserved in the SF-UIDA framework.\par
	\item 
	\textbf{SO+SSA+SSP}: we employed the full two-step alignment process of the SF-UIDA framework.\par
\end{itemize}
%As shown in Tab.\ref{table:ablation}, the results of our ablation experiments have convincingly demonstrated the efficacy of our proposed two-step alignment methodology tailored for unsupervised individual domain adaptation. When compared to the source only method, both the SSA and SSP modules can achieve better performances, demonstrating the effectiveness of our proposed two-step alignment strategy during process of the model customization. Specifically, the SSP module performs slightly better than the SSA module. This may imply that during the process of individual domain adaptation, aligning the class-conditional distributions of the individual target domain contributes more significantly to adaptation than aligning the target domain's marginal distribution. It is reasonable that the sleep data is imbalanced in terms of classes, and we can observe a larger gap between the SSP and SSA modules on the MF1 metric (67.7\% vs. 66.8\%) compared to the ACC metric (75.0\% vs. 74.8\%), as the MF1 metric can better reflect the accuracy of correct classification for each class. By combining the SSA and SSP alignment modules, our SF-UIDA framework achieves better performance than the single alignment based approach, enabling the personalized customization of the source model.
As shown in Table \ref{table:ablation}, our ablation experiments convincingly demonstrate the efficacy of the proposed two-step alignment methodology for unsupervised individual domain adaptation. Compared to the source-only method, both the SSA and SSP modules yield improved performance, highlighting the effectiveness of our alignment strategy during model customization. Notably, the SSP module outperforms the SSA module slightly, suggesting that aligning the class-conditional distributions of the individual target domain is more impactful than aligning the marginal distribution. This is reasonable given the imbalance in sleep data classes, as evidenced by a greater performance gap on the MF1 metric (67.7\% vs. 66.8\%) compared to the ACC metric (75.0\% vs. 74.8\%). The MF1 metric more accurately reflects the classification performance for each class. By integrating both SSA and SSP alignment modules, our SF-UIDA framework achieves superior performance compared to single-alignment approaches. 
%facilitating personalized customization of the source model.
\begin{figure}[!tb]
	\centering
	\includegraphics[width=0.95\columnwidth]{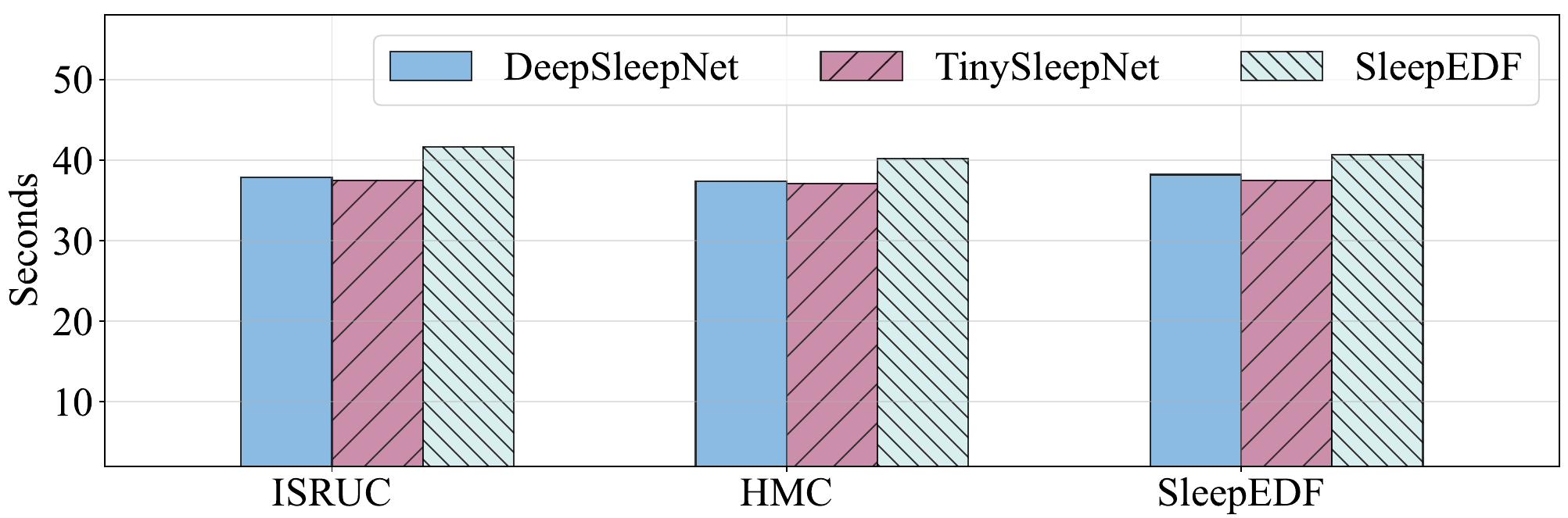}
	\caption{Average time cost per individual (seconds). }
	\label{fig:7}
\end{figure}
\subsubsection{Computational Complexity}
To evaluate the computational complexity of our proposed SF-UIDA framework, we calculate the time cost per individual across three datasets shown in Fig. \ref{fig:7}. Our method is capable of completing personalized  customization for an unknown individual within an average of 40 seconds. When compared to the several hours' duration of one person's sleep records, this time cost is acceptable. Moreover, considering the unseen subject commonly appears one by one in practice, our method is applicable to enable the source model continuously adapt to new subjects and achieve plug-and-play personalized customization for each individual.
\begin{figure}[!tb]
	\centering
	\includegraphics[width=1.0\columnwidth]{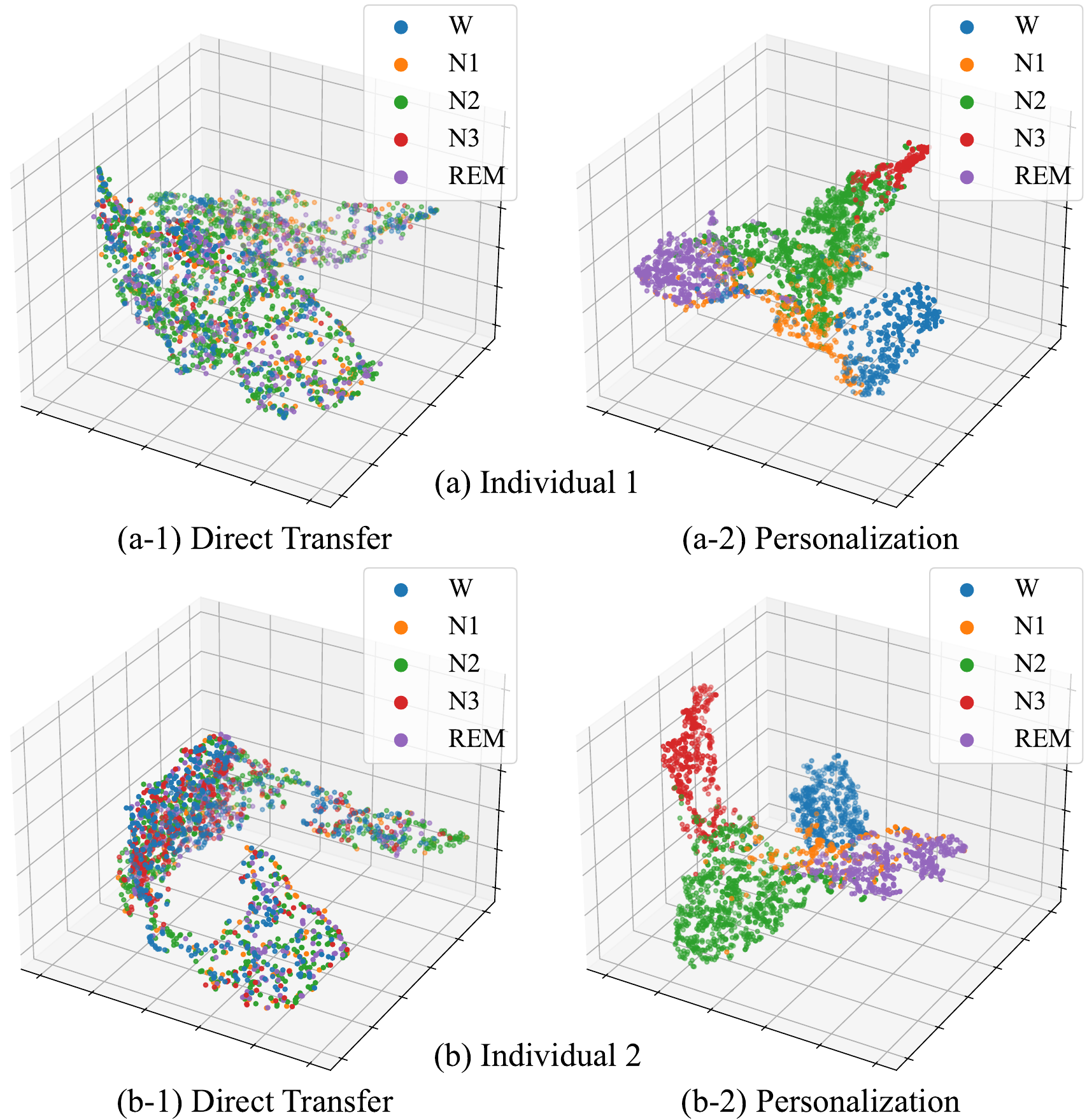}
	\caption{Sleep features visualization.}
	\label{fig:6}
\end{figure}
\subsubsection{Feature Visualization}
%To demonstrate the effectiveness of our method, we selected two individuals from the Sleep-EDF dataset to visualize the intermediate features based on the t-SNE method \cite{van2008visualizing}. Fig.\ref{fig:6} (a-1) and Fig.\ref{fig:6} (a-2) illustrate the feature distribution by direct inference and the distribution after personalization for the first individual. Fig.\ref{fig:6} (b-1) and Fig.\ref{fig:6} (b-2) depict the distribution for the second individual. As clearly depicted by the visualized distribution of the first individual, the samples belonging to the same sleep stage are nicely clustered within the same cluster in Fig.\ref{fig:6} (a-2), compared with the samples in Fig.\ref{fig:6} (a-1). The phenomenon is similar for the second individual. This demonstrates that our method can effectively achieve personalized customization for target subjects.
To demonstrate the effectiveness of our method, we selected two individuals from the Sleep-EDF dataset to visualize intermediate features using the t-SNE method \cite{van2008visualizing}. Figures \ref{fig:6} (a-1) and (a-2) illustrate the feature distribution for the first individual, showing the distribution before and after personalization. Similarly, Figures \ref{fig:6} (b-1) and (b-2) depict the distribution for the second individual. As shown in Figure \ref{fig:6} (a-2), samples from the same sleep stage are well-clustered after personalization, compared to the distribution in Figure \ref{fig:6} (a-1). A similar trend is observed for the second individual, demonstrating that our method effectively achieves personalized customization for target subjects.
\section{CONCLUSIONS}
%In this paper, we proposed a novel Source-Free Unsupervised Individual Domain Adaptation (SF-UIDA) framework for automatic sleep staging, which adopts a two-step subject-specific alignment scheme for adaptation. Our framework enables the personalized customization to be plug-and-play applied to each newly appeared individual without access to the source data, meeting the practical needs in clinics. The experimental results on three public datasets demonstrate that our SF-UIDA framework can effectively transform a source model into a personalized model within a short time adaptation, which holds significance in practice. However, there still exists limitations. Our framework is specially designed for the sleep staging task which may not generalize well on other downstream EEG tasks. In future work, we intent to expand the application of our method to a wider range of EEG tasks, especially for tasks based on wearable devices, rather than limiting it to sleep staging.
In this paper, we present a novel Source-Free Unsupervised Individual Domain Adaptation (SF-UIDA) framework for automatic sleep staging, employing a two-step subject-specific alignment scheme for adaptation. Our framework facilitates plug-and-play personalization for each new individual without requiring access to source data, meeting the practical needs in clinics. Experimental results across three public datasets demonstrate that the SF-UIDA framework effectively transforms a source model into a personalized one within a short adaptation period, highlighting its practical significance. Our future work will aim to extend the applicability of our method to a broader range of EEG tasks.

%However, limitations exist: the framework is specifically designed for sleep staging and may not generalize well to other EEG tasks. Future work will aim to extend the applicability of our method to a broader range of EEG tasks, particularly those involving wearable devices, beyond sleep staging.

\section{Acknowledgments}
This work was supported by STI 2030 Major Projects (2021ZD0200400), the Key Program of the Natural Science Foundation of Zhejiang Province, China (No. LZ24F020004), the Natural Science Foundation of China (No. 61925603) and the Fundamental Research Funds for the Central Universities (No. 61925603). The corresponding author is Dr. Sha Zhao.

\bibliography{aaai25}

\end{document}